\documentclass[11pt]{article}
\usepackage[numbers, compress]{natbib}
\usepackage[margin=1in]{geometry}
\usepackage{subcaption}

\usepackage[table]{xcolor}  
\definecolor{Gray}{gray}{0.9}
\definecolor{midgreen}{rgb}{0.1,0.5,0.1}
\definecolor{darkgray}{gray}{0.25}
\definecolor{lightblue}{rgb}{0.25,0.25,0.8}
\definecolor{mydarkblue}{rgb}{0,0.08,0.45}
\usepackage[colorlinks,
linkcolor=lightblue,
anchorcolor=blue,
urlcolor=mydarkblue,
citecolor=lightblue]{hyperref}
\usepackage{makebox}
\usepackage{amsmath}
\usepackage{amssymb}
\usepackage{mathtools}
\usepackage{amsthm}
\usepackage{setspace}
\usepackage{bm}
\usepackage{bbm}

\usepackage{microtype}
\usepackage{hyperref}
\usepackage{amsmath, amsthm,amssymb}

\usepackage{url}
\usepackage{booktabs}

\usepackage{lineno}
\usepackage{xcolor}
\definecolor{modernblue}{rgb}{0.3, 0.4, 0.6}
\usepackage[most]{tcolorbox}
\usepackage{mdframed}
\usepackage{tikz}
\usepackage{fontawesome5}
\usepackage{wrapfig}

\usetikzlibrary{positioning,shadows.blur,backgrounds, fit}

\definecolor{darkblue}{rgb}{0, 0, 0.5}
\hypersetup{colorlinks=true, citecolor=darkblue, linkcolor=darkblue, urlcolor=darkblue}

\title{Brains vs. Bytes: Evaluating LLM Proficiency in Olympiad Mathematics}

\begin{document}


\author{
  Hamed Mahdavi \\
  Pennsylvania State University \\
  \texttt{hmm5834@psu.edu}
  \and
    Alireza Hashemi \\
  City University of New york \\
  \texttt{alireza.hashemi13@outlook.com}
  \and
  Majid Daliri \\
  New York University \\
  \texttt{daliri.majid@nyu.edu}
  \and
  Pegah Mohammadipour \\
  Pennsylvania State University \\
  \texttt{pegahmp@psu.edu}
  \and
  Alireza Farhadi \\
  Amirkabir University of Technology  \\
  \texttt{farhadi@aut.ac.ir}
  \and
  Samira Malek \\
  Pennsylvania State University \\
  \texttt{sxm6547@psu.edu}
  \and
  Yekta Yazdanifard \\
  Bocconi University \\
  \texttt{Yekta.yazdanifard@unibocconi.it}
  \and
  Amir Khasahmadi \\
  Autodesk \\
  \texttt{amir.khasahmadi@autodesk.com}
  \and
  Vasant Honavar \\
  Pennsylvania State University \\
  \texttt{vhonavar@psu.edu}
}

\maketitle
\begin{abstract}
Recent advances in large language models (LLMs) have shown impressive progress in mathematical reasoning tasks. However, current evaluation benchmarks predominantly focus on the accuracy of final answers, often overlooking the crucial logical rigor for mathematical problem solving. The claim that state-of-the-art LLMs can solve Math Olympiad-level problems requires closer examination. To explore this, we conducted both qualitative and quantitative human evaluations of proofs generated by LLMs, and developed a schema for automatically assessing their reasoning capabilities. Our study reveals that current LLMs fall significantly short of solving challenging Olympiad-level problems and frequently fail to distinguish correct mathematical reasoning from clearly flawed solutions. Our analyses demonstrate that the occasional correct final answers provided by LLMs often result from pattern recognition or heuristic shortcuts rather than genuine mathematical reasoning. These findings underscore the substantial gap between LLM performance and human expertise in advanced mathematical reasoning and highlight the importance of developing benchmarks that prioritize the soundness of the reasoning used to arrive at an answer rather than the mere correctness of the final answers.
\end{abstract}

\section{Introduction}

The release of OpenAI's o1 model \citep{openai2024openaio1card} marks a significant breakthrough in artificial intelligence research, particularly in the domains of reasoning and problem solving. Building on this achievement, several state-of-the-art models have been introduced \citep{deepseekai2024deepseekv3technicalreport,qwen2025qwen25technicalreport,google_gemini_v2}, which incorporate post-training on chain-of-thought data. These models have been claimed to demonstrate enhanced reasoning abilities in solving mathematical problems. Although the development of reasoning models remains an active area of research, post-training techniques are believed to be pivotal for improving performance on tasks requiring planning, iterative thinking, and trial-and-error strategies. It has been suggested that generating reasoning tokens before producing a final answer improves the reliability of solutions produced by these models for complex reasoning tasks, e.g., for mathematical problem solving.

A range of different benchmarks, such as GSM8K \citep{cobbe2021gsm8k} and MATH \citep{hendrycks2021measuringmathematicalproblemsolving}, have been used to assess the mathematical reasoning abilities of large language models (LLM). With these models achieving competitive performance on such benchmarks, for example, Qwen2.5-72B is reported to reach 91. 5\% accuracy on the GSM8K benchmark and 62.1\% on the MATH benchmark, and OpenAI O1 94.8\% on the MATH benchmark, there have been efforts aimed at introducing more challenging benchmarks. These include OlympiadBench \citep{he2024olympiadbench}, OlympicArena \citep{huang2024olympicarena}, CHAMP \citep{mao-etal-2024-champ}, AlphaGeometry \citep{AlphaGeometryTrinh2024}, MathOdyssey \citep{fang2024mathodysseybenchmarkingmathematicalproblemsolving}, and Omni-MATH \citep{gao2024omnimathuniversalolympiadlevel}, which contain mathematical problems at the contest level designed to challenge the mathematical problem solving capabilities of LLMs.

Except for CHAMP, which includes "concepts, general math facts, and hints" \citep{mao-etal-2024-champ} as additional annotations, and AlphaGeometry, which contains problems and solutions expressed in mathematical language, all other benchmarks use only the correctness of the symbolic or numeric answer correctness, as the evaluation metric to assess the reasoning capabilities of various language models. Because they do not adequately assess the correctness of the rationale or reasoning used to produce the answer, they fail to assess the reasoning capabilities of the LLMs being evaluated.

Thus, it is possible for models to use flawed heuristics or erroneous reasoning to score well on the assessment as long as they happen to somehow arrive at the correct final answer. In the case of widely used benchmarks, such as GSM8K and MATH, which consist mostly of simpler problems, where arriving at the correct answer is indistinguishable from coming up with a provably correct solution. However, this is not the case with benchmarks like CHAMP, OlympiadBench, Omni-MATH, MathOdyssey, and OlympicArena, which include contest-level problems where it is sometimes possible to guess a correct answer without being able to offer a sound rationale (or proof) that the answer is correct. This chasm between producing a correct answer as opposed to producing a correct answer backed up by a sound rationale to justify the anser becomes particularly crucial when the LLM is presented with open-ended questions.

Due to the cost and labor involved in assessing the accuracy of the reasoning used by LLM in solving mathematical problems, there is growing interest in LLM-as-a-Judge, a technique to use an LLM to automatically assess the quality of LLM-generated responses \citep{li2025generationjudgmentopportunitieschallenges, li2024llmsasjudgescomprehensivesurveyllmbased, cohen2023lm, adlakha-etal-2024-evaluating, liu-etal-2023-g, manakul2023selfcheckgpt}. We note that direct evaluation of LLM-generated solutions, as in \cite{dubois2023alpacafarm}, can be rather complex for challenging math problems, due to the effort involved in verifying the correctness of an LLM-generated solution and the rationale used to justify it. The application of consistent rubrics for automated scoring of LLM-generated solutions is also nontrivial because of the variability of the relative importance of different facts across problems and their solutions. In the case of math Olympiad problems, we find that some mistakes show a lack of understanding of some fundamental concepts, whereas others may be more superficial and amenable to being rectified with minor adjustments. There are no clear and objective criteria to distinguish between these two categories of errors. \cite{sawada2023arb} proposes a rubric-based approach in which a reference solution and specified evaluation criteria are used to assess the quality of a generated solution. However, this approach faces a serious limitation: There may be correct solutions that may not align with the given reference solution. 

\begin{wrapfigure}{r}{0.31\textwidth}
  \centering
  \includegraphics[width=0.31\textwidth]{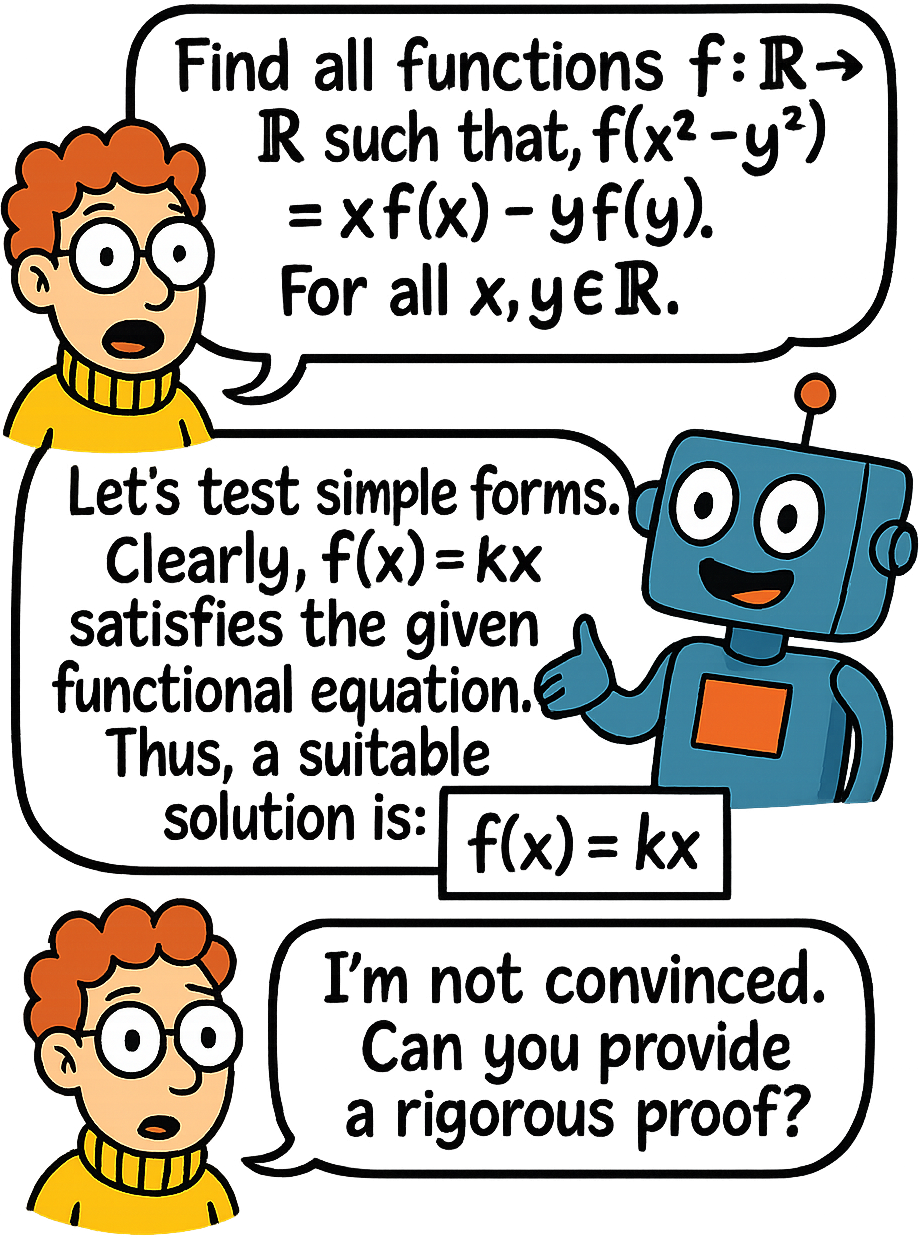}
  \caption{ LLMs often fail to generate logically sound solutions for challenging problems.}
  \vspace{-10pt}

  \label{fig:wrapped}
\end{wrapfigure}

Given these limitations, our goals in this work are twofold: first, to rigorously assess the quality of LLM-generated solutions to mathematics Olympiad problems, including identifying their common failure modes and error types; and second, to investigate whether LLMs can verify the correctness of LLM-generated solutions. This study is significant for two main reasons. First, evaluating the correctness of the solutions generated by LLMs can reveal whether training strategies that rely solely on the correctness of the final answer (as opposed to the correctness of the reasoning used to justify the answer) are sufficient to guide LLMs toward producing correct solutions backed up by sound reasoning. Second, evaluating the ability of LLMs to verify their own generated solutions can provide insights into the feasibility of bootstrapping verification processes, potentially enabling LLMs to improve the quality of their solutions.

Toward these goals, we assembled a group of evaluators to assess the quality of LLM-generated solutions for problems from the International Mathematics Olympiad (IMO) shortlist. After carefully analyzing solutions generated by frontier models such as o1, o1-mini, o3-mini, Gemini 2.0 (Flash Thinking mode) and DeepSeek R1\footnote{At the time we began this project, reasoner models such as Claude 3.7, o3-mini, Gemini 2.5 and Grok 3 had not yet been released; Among these, we were only able to include o3-mini in our evaluations.}, we observed that only a negligible fraction of LLM-generated solutions were correct. We found that LLM-generated solutions frequently contained common types of fallacies that invalidated their reasoning. We grouped the observed fallacies into several categories.  We systematically evaluated the solutions generated by the frontier models with respect to their correctness. Furthermore, we labeled each incorrect solution according to the type of fallacy they contained. We also evaluated the ability of the frontier models to verify LLM-generated solutions. The results of our analyses show that, in nearly all cases, the models failed to distinguish between correct solutions and incorrect ones containing fallacious reasoning steps. In light of the fact that identifying the presence of fallacies in reasoning is a simpler task compared to full verification of the solution,  we conclude that LLMs are incapable of verifying the correctness of LLM-generated solutions. In summary, our contributions are as follows. 

\begin{itemize}
    \item We conducted an extensive evaluation of solutions generated by frontier models on 455 IMO shortlist problems, with emphasis on the correctness of reasoning used to arrive at the answer, in addition to the correctness of the answer itself.
    \item To gain deeper insights into the recurring mistakes made by LLMs, we systematically identified and categorized common logical fallacies present in LLM-generated solutions, offering a comprehensive framework for classifying typical errors.
    \item Based on the insights offered by the preceding findings, we created a labeled data set by annotating each solution according to the correctness and the type(s) of fallacy or fallacies it contains. This data set supports tasks related to solution verification and offers valuable insights into the current capabilities and limitations of the frontier LLMs.
    \item Our analysis reveals that even advanced models frequently struggle to distinguish between valid solutions and those containing evident logical fallacies.
\end{itemize}

\section{Related Work}

\textbf{Benchmarks:} Several benchmarks have been used to assess the mathematical reasoning capabilities of large language models (LLMs) \citep{ahn2024largelanguagemodelsmathematical}. Some benchmarks focus purely on arithmetic problems \citep{yuan2023largelanguagemodelsperform}, while data sets for math word problems (MWP), such as GSM8K \citep{cobbe2021gsm8k} and MathQA \citep{amini2019mathqainterpretablemathword}, present natural language scenarios requiring logical reasoning \citep{wei2023chainofthoughtpromptingelicitsreasoning}. Automated theorem proof (ATP) data sets evaluate models' capabilities in logical theorem proving \citep{zheng2022minif2fcrosssystembenchmarkformal, yu2024metamathbootstrapmathematicalquestions, jiang2024leanreasonerboostingcomplexlogical}. Recent benchmarks focus on advanced or Olympiad-level mathematics, such as CONIC10K for conic sections \citep{wu2023conic10kchallengingmathproblem}, GHOSTS and miniGHOSTS for graduate-level mathematics \citep{frieder2023mathematicalcapabilitieschatgpt}, and CHAMP \citep{mao-etal-2024-champ}, OlympiadBench \citep{he2024olympiadbench}, MathOdyssey \citep{fang2024mathodysseybenchmarkingmathematicalproblemsolving}, and Omni-MATH \citep{gao2024omnimathuniversalolympiadlevel}, specifically focusing on competition-level problems. HARP \citep{yue2024harpchallenginghumanannotatedmath} provides human-annotated US competition problems, and NuminaMath offers a large-scale collection of math problems and solutions \citep{numina_math_datasets}.

\textbf{LLM-as-a-judge:} Some have advocated the use of LLMs as judges to assess the mathematical reasoning abilities of LLMs, to minimize the need for expensive human annotations \citep{stephan2024calculationadjudicationexaminingllm, li2024llmsasjudgescomprehensivesurveyllmbased, nasrabadi2024jureejudgessafeguardingllm, ning2024picopeerreviewllms}. This paradigm offers adaptable evaluations based on task-specific contexts \citep{tan2024largelanguagemodelsdata, dhurandhar2024rankinglargelanguagemodels}, and its effectiveness is typically measured against human judgments \citep{kim2024demonstrationadaptivecollaborationlarge, ye2024justiceprejudicequantifyingbiases, liu2025aligninghumanjudgementrole}. Benchmarks such as UltraFeedback \citep{cui2024ultrafeedbackboostinglanguagemodels}, AlpacaEval \citep{dubois2024lengthcontrolledalpacaevalsimpleway}, Chatbot Arena \citep{chiang2024chatbotarenaopenplatform}, and MT-Bench \citep{zheng2023judgingllmasajudgemtbenchchatbot} evaluate different LLM judging domains. Specifically for mathematical reasoning, REASONEVAL \citep{xia2025evaluatingmathematicalreasoningaccuracy} assesses the correctness of the answer and of the reasoning used to arrive at the answer, while MATHCHECK \citep{zhou2024modelreallygoodmath} uses LLM to assess the performance of LLM in a variety of mathematical problems. The SMART-840 dataset \citep{cherian2024evaluatinglargevisionandlanguagemodels} benchmarks zero-shot mathematical reasoning based on human performance statistics.

\textbf{Mathematical Reasoning in LLMs:} Large language models (LLMs) have shown success in various reasoning tasks, especially when employing prompting techniques like Chain-of-Thought (CoT), which encourages them to generate correct intermediate steps toward a solution \citep{chen2024languagemodelshiddenreasoners,wei2023chainofthoughtpromptingelicitsreasoning,kojima2023largelanguagemodelszeroshot}. These methods can significantly improve performance on challenging problems \citep{havrilla2024glorewhenwhereimprove}. Furthermore, inference-time techniques such as CoT with Self-Consistency (CoT-SC) have been developed to improve reasoning by generating multiple reasoning paths and selecting the most consistent one \citep{wang2023selfconsistencyimproveschainthought,wang2024chainofthoughtreasoningprompting}. Benchmarks such as MATH \citep{hendrycks2021measuringmathematicalproblemsolving}, GSM-Symbolic, and GSM-NoOp \citep{mirzadeh2024gsmsymbolicunderstandinglimitationsmathematical}  have been introduced to provide more controllable evaluations and reveal limitations such as sensitivity to numerical variations and irrelevant information, suggesting a potential lack of deep understanding of mathematical concepts. These benchmarks show that current LLMs rely more on probabilistic pattern matching than genuine formal logical reasoning. To further refine LLMs' reasoning, approaches like reward modeling to evaluate solution correctness and self-refinement techniques \citep{huang2024largelanguagemodelsselfcorrect} and decomposing problems into smaller algorithmic steps \citep{zelikman2023parselalgorithmicreasoninglanguage} are being explored.

\section{General Workflow}

In this section, we outline the data collection process. In the first phase, we selected a set of challenging problems to evaluate the quality of solutions generated by the LLMs. We gathered a group of seven evaluators, each either a former national-level Olympiad medalist or holding or doing a relevant Ph.D. in fields like mathematics or computer science, and asked them to analyze the correctness of the LLM-generated solutions. 
\subsection{Problem Selection Rationale}
 Through this project, we used IMO shortlist problems. The process of selecting problems for the IMO Shortlist is rigorous and carefully coordinated. Each participating country submits a set of candidate problems that typically span the four main mathematical fields: algebra, geometry, combinatorics, and number theory. These submissions are reviewed by a problem selection committee to ensure that they meet key criteria, including originality, mathematical depth, and suitability for the competition. The committee carefully evaluates the problems for their difficulty level, ensuring a balance between accessibility for less experienced participants and sufficient challenge for the most advanced contestants. From this review process, a shortlist is created that contains a diverse collection of high-quality problems. This shortlist forms the basis for the final selection of problems used in the IMO. 

The shortlist problems have some distinct features that make them suitable for testing the mathematical reasoning capabilities of the frontier models:

\begin{itemize} 
\item Shortlist problems are highly original, even within the context of contest-level problems. The selection committee ensures that the solution ideas for these problems are as novel and unique as possible. As a result, while attempting to generate solutions, an LLM cannot simply combine standard building blocks from well-known problems to arrive at the correct answer. 
\item It is almost always known that the problem can be reduced to a well-established result in advanced mathematics, such as undergraduate- or graduate-level topics or research-level findings. Consequently, an LLM cannot leverage its extensive knowledge to apply an advanced mathematical result for an easy solution. 
\item All problems are designed to be solvable using high-school-level mathematics, and the challenge lies in the intricacy of the ideas rather than in requiring a background in advanced mathematics. 
\item The solutions typically involve multiple steps, each requiring nuanced arguments. Solving these problems requires careful planning, systematic thinking, and rigorous verification of each step, in contrast to simpler mathematical problems that can be tackled through straightforward algebraic manipulations or trial-and-error. \end{itemize}

On the other hand, since the IMO and IMO Shortlist problems are highly reputable, there is a significant likelihood of data leakage, as frontier models may have been trained on publicly available high-quality mathematical datasets. In this paper, we demonstrate that even if such data leakage has occurred, it does not substantially affect the ability of the LLM to solve IMO Shortlist-level problems. 

\subsection{Classifying the Failure Modes of LLM Solutions}
We asked the evaluators to present several IMO shortlist and shortlist-level problems to frontier models, including OpenAI o1, o1-mini, o3-mini, DeepSeek R1 and Gemini 2, and qualitatively analyze the details of the LLM-generated arguments. We found that when these frontier models generate incorrect solutions, errors consistently follow common patterns. Specifically, incorrect solutions typically involve blatantly inaccurate mathematical arguments or statements.

After thorough analysis, we identified the following types of reasoning fallacies in incorrect LLM-generated solutions. Demonstrative and real examples of each type of reasoning fallacy are provided in the appendix.

\textbf{Proof by Example.}  
Drawing a general conclusion based on a limited number of specific instances without rigorous justification for all cases. This error occurs when a mathematical claim appears to hold in a few examples, misleadingly suggesting that it is universally true when, in fact, it is not.

\textbf{Proposal Without Verification.}  
Introducing a method or strategy without properly justifying its correctness. The model proposes an idea but provides no rigorous argument or proof supporting its validity.

\textbf{Inventing Wrong Facts.}  
Citing or inventing non-existent theorems, definitions, or facts to justify a claim. Instead of relying on established mathematical facts, the argument relies on fabricated statements (hallucination).

\textbf{Begging the Question (Circular Reasoning).}  
Assuming the conclusion it that needs to be proved, instead of providing evidence for the claim.

\textbf{Solution by Trial-and-Error.}  
Offering solutions derived solely from guesswork or testing a few random examples without providing a reason as to why selected solutions work or why alternatives are not considered.

\textbf{Calculation Mistakes.}  
Committing substantial arithmetic or algebraic errors that undermine the overall correctness of the solution. We specifically considered calculation errors severe enough to compromise the validity of the conclusion.

\subsection{Data Annotation Processs}
After defining the fallacy categories, we provided a list of IMO shortlist problems and the corresponding model-generated solutions to the evaluators, instructing them to classify these solutions based on their correctness. We used the following checklist to annotate our data:

\begin{enumerate}

\item First, the evaluator read the solution and determined whether it was correct, partially correct, or incorrect. 

\begin{itemize}
    \item A solution was considered \textit{correct} if it fully addressed all aspects of the problem and contained no significant errors in statements or conclusions. 
    \item A solution was deemed \textit{partially correct} if it included some essential steps of a correct solution but omitted other crucial steps or contained significant inaccuracies. 
    \item A solution was classified as \textit{incorrect} if it lacked any useful non-trivial information relevant to solving the problem. 

\end{itemize}
\item If the solution was not correct, the evaluator identified the fallacy (or fallacies) that occur in the solution. 

\item If the problem required a final answer, the evaluator recorded both the correct final answer and the final answer generated by the model.

\end{enumerate}

To ensure consistency, the evaluators' team lead conducted a thorough review of the evaluators' outputs in parallel, verifying that the definitions of fallacies were applied correctly and consistently across all evaluators. Borderline cases were identified and discussed separately.

\section{Human Evaluation Results}
We evaluated the models using the sets of IMO shortlist problem from 2009 to 2023, comprising a total of 455 problems. These included 108 algebra, 117 combinatorics, 116 geometry, and 114 number theory problems. The number of problems per annual shortlist varied slightly, typically ranging from 26 to 35. Each set was carefully curated to maintain a balanced distribution of difficulty on the four primary mathematical topics. The performance of each model on the IMO shortlist problems is summarized in Table \ref{tab:llm_accuracy}. As evident from the results, none of these models achieve performance levels comparable to those obtained by calculating the accuracy of the final answers, as reported in \cite{gao2024omnimathuniversalolympiadlevel} and \cite{fang2024mathodysseybenchmarkingmathematicalproblemsolving}.

\begin{table}[ht]
\centering
\small
\begin{tabular}{lccc}
\toprule
\textbf{Model} & \textbf{Correct (\%)} & \textbf{Partially Correct (\%)} & \textbf{Incorrect (\%)} \\
\midrule
DeepSeek & 3.8 & 6.7 & 89.4 \\
Gemini 2.0 & 0.0 & 1.1 & 98.9 \\
o1 & 1.9 & 3.9 & 94.2 \\
o1-mini & 0.0 & 0.0 & 100.0 \\
o3-mini & 3.3 & 4.4 & 92.2 \\
\bottomrule
\end{tabular}
\caption{Performance of different models on IMO shortlist problems (\%)}
\label{tab:llm_accuracy}
\end{table}

The observed gap between the outcomes of our evaluation and other methods that focus exclusively on the correctness of the final can be explained by the fact that it is possible for a model to get the final answer right based on fallacious reasoning steps. To investigate this issue, we specifically examined problems with concrete final answers within our evaluation set. Table \ref{tab:final_answers} illustrates both the proportion of correct final answers and the conditional probability of having a correct solution, given that the final answer is correct. Interestingly, we found that the frontier models often generate incorrect solutions, that is, solutions that contain reasoning errors despite getting the final answer right.

\begin{table}[htbp]
\centering
\renewcommand{\arraystretch}{1.2}
\setlength{\tabcolsep}{8pt}
\begin{tabular}{lcc}
\toprule
\textbf{Model} & \textbf{Final Answer Accuracy (\%)} & \textbf{Correct$|$Correct Final Answer (\%)} \\
\midrule
DeepSeek & 63.2\% & \phantom{0}0\% \\
Gemini 2.0 & 43.8\% & \phantom{0}0\% \\
o1 & 30.8\% & 12.5\% \\
o1-mini & 35.0\% & \phantom{0}0\% \\
o3-mini & 48.3\% & 14.3\% \\
\bottomrule
\end{tabular}
\caption{Comparison of evaluated LLMs highlighting the gap between final answer correctness and overall solution quality. \textbf{Final Answer Accuracy} denotes the percentage of correct final answers, whereas \textbf{Correct$|$Correct Final Answer} represents the percentage of fully correct solutions among instances where the final answer is correct.}
\label{tab:final_answers}
\end{table}

These results suggest that the models use heuristics, shortcuts, and educated guesses that often turn out to be misguided instead of using sound reasoning. Consequently, evaluation of such models solely on the basis of the correctness of the final answer tends to grossly overestimate their reasoning abilities. Hence, we argue that evaluating the performance of LLMs on mathematical reasoning tasks based solely on the correctness of the final answer without checking the soundness of the rationale used to justify the answer is fundamentally flawed. 
\begin{figure}[h]
    \centering
    \includegraphics[width=0.95\linewidth]{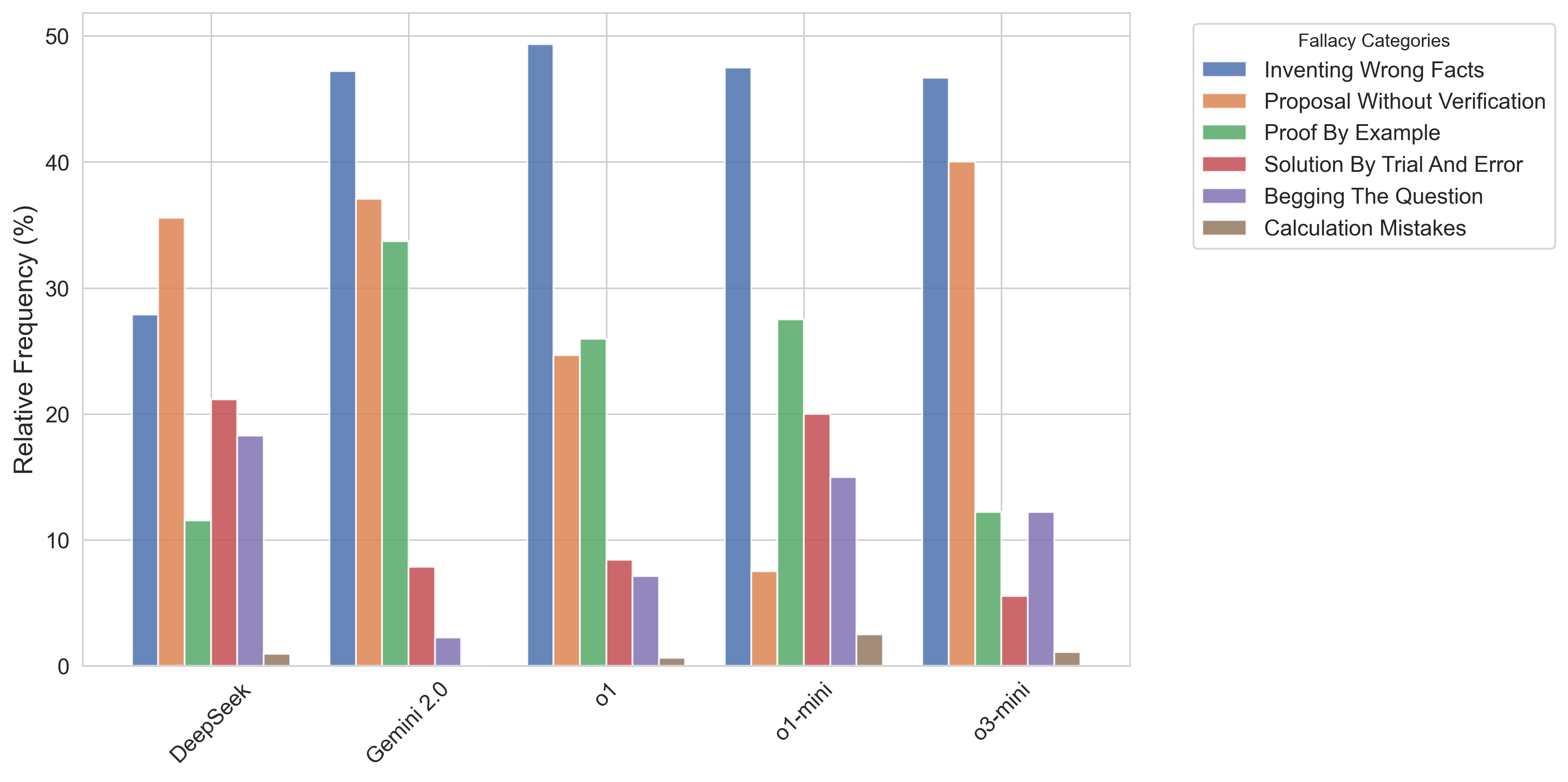}
    \caption{Relative frequencies of each fallacy among the LLM-generated solutions for each model. }
    \label{fig:relative_frequencies}
\end{figure}

It is also helpful to examine how the reliance of common reasoning fallacies varies across the different models. Figure \ref{fig:relative_frequencies} shows the relative frequency of occurrence of each type of fallacy in the LLM-generated solutions produced by each of the models. Note that some solutions can include multiple types of fallacies. 

We observe that among the different types of fallacies, \textit{Inventing Wrong Facts} is the most frequent in four of the models and the second most frequent in a fifth model. We conjecture that this observation can be explained by the fact that these models are trained with reinforcement learning using a reward function based on the correctness of the final answers (e.g., see the DeepSeek report; \citep{deepseekai2024deepseekv3technicalreport}).

The results show that \textit{Proposal Without Verification} is the second most frequently observed reasoning fallacy. Although the internal reasoning tokens are not visible for OpenAI models, an examination of the "thinking traces" from DeepSeek and Gemini suggests that this fallacy often arises because the model struggles to decide which calculations and statements to include in its response. Consequently, instead of presenting well-reasoned rationale or useful intermediate steps, the model may produce vague claims, often beginning with phrases like "It is easy to show that..." or "One can show that..." without justification or supporting arguments.

\begin{wrapfigure}{r}{0.60\textwidth}
  \centering
  \includegraphics[width=0.60\textwidth]{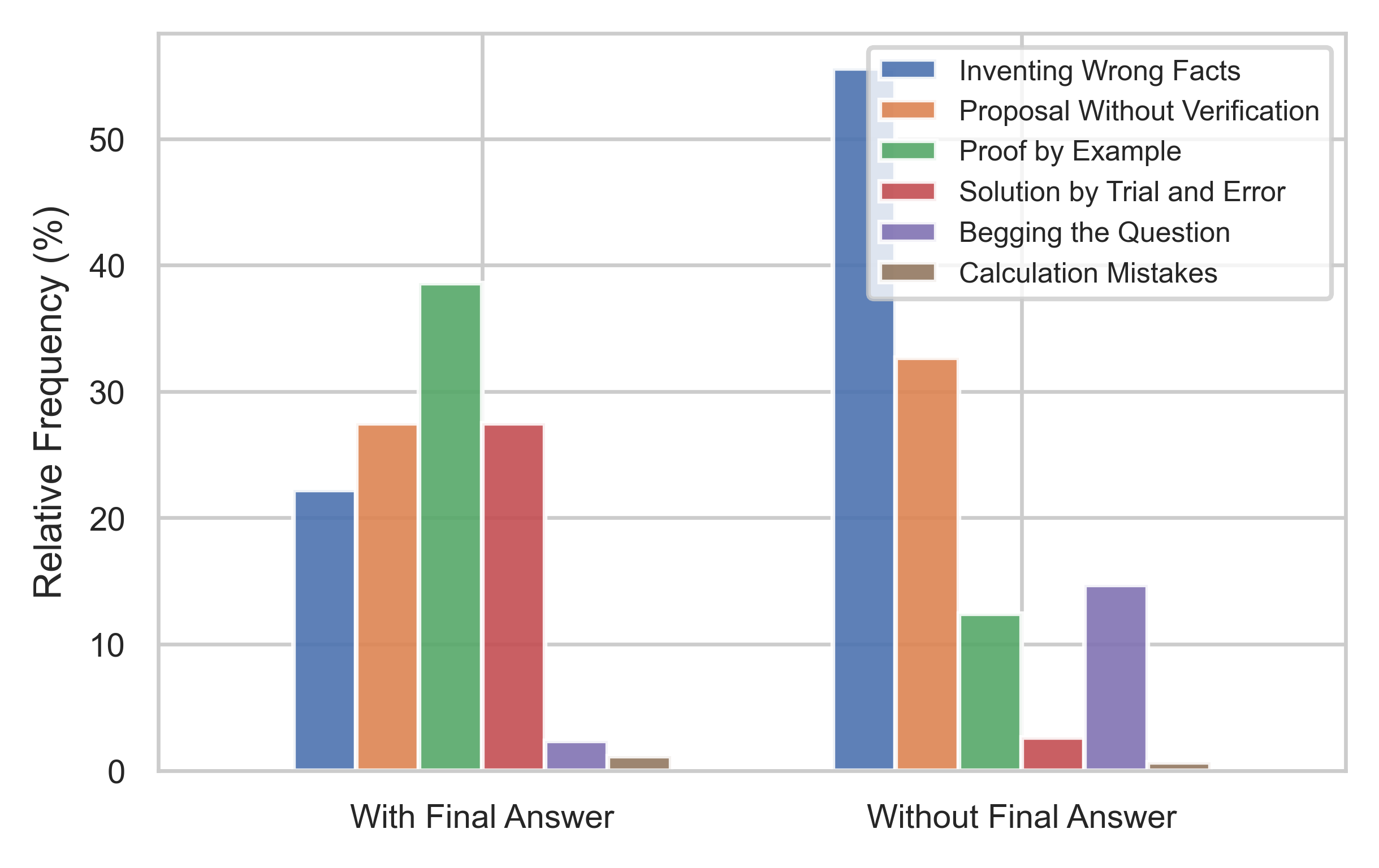}
  \caption{Relative frequencies of each fallacy in LLM-generated solutions, comparing questions with and without a final answer.}
  \label{fig:fallacy-frequencies-final-answers}
  \vspace{-8pt}
\end{wrapfigure}

We also observed that models tend to exhibit different types of fallacious reasoning depending on the type of problem presented to them. Figure \ref{fig:fallacy-frequencies-final-answers} illustrates the relative frequencies of various fallacies depending on whether or not the solutions produced contain final answers. Notably, \textit{Proof by Example} and \textit{Solution by Trial and Error} occur more frequently in solutions that include a final answer. This suggests that models often arrive at final answers through either heuristic trial and error or generalization from a small number of test cases, leading to a higher prevalence of these two types of fallacy in the solutions produced. In contrast, we observe that \textit{Inventing Wrong Facts} and \textit{Proposal Without Verification} occur more frequently in solutions of problems that lack explicit final answers. To correctly solve such problems, the models must logically connect the problem's initial assumptions and constraints to the conclusion through a sequence of sound reasoning steps. However, we find that models frequently substitute sound reasoning with unjustified and often incorrect statements.

\begin{figure}[t]
    \centering
    \includegraphics[width=0.95\linewidth]{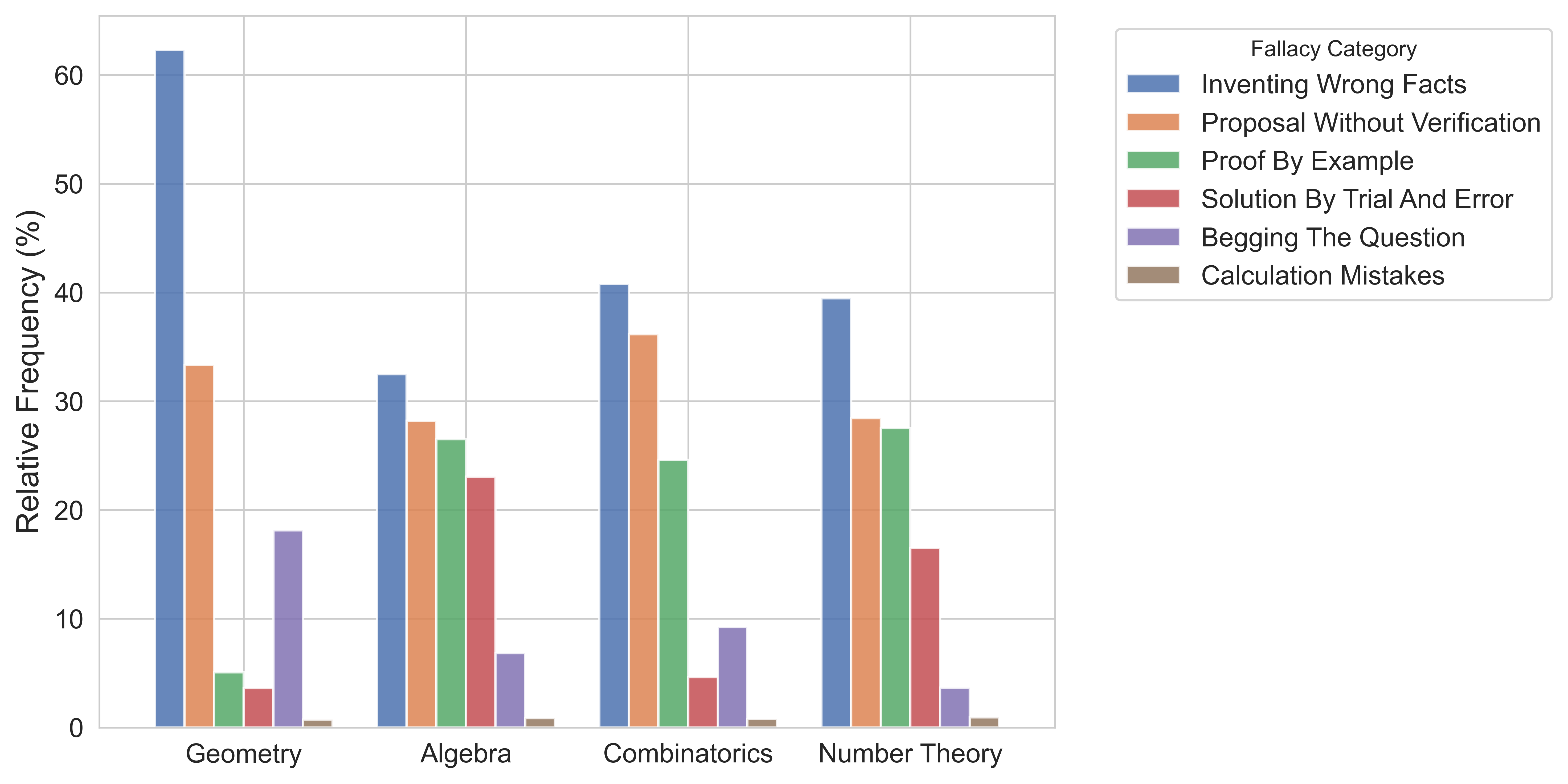}
    \caption{Relative frequencies of each fallacy in LLM-generated solutions among different topics}
    \label{fig:fallacy-frequencies-topics}
\end{figure}

We also observe differences in the relative frequencies of the fallacies in problems drawn from different topics in mathematics. Figure \ref{fig:fallacy-frequencies-topics} demonstrates the relative frequencies of reasoning fallacies among geometry, algebra, combinatorics,  and number theory problems. As more geometry problems can be solved only using logical statements rather than algebraic manipulations, \textit{Inventing Wrong Facts}, \textit{Proposal Without Verification}, and \textit{Begging the Question} are more common in geometry problems. A significant proportion of algebra problems fall into categories such as functional equations, polynomial equations, or optimization tasks. We observed that all frontier models tend to avoid generating rigorous analytic solutions, instead relying on trial and error to determine the final answer. This behavior results in a higher frequency of the \textit{Solution by Trial and Error} fallacy in LLM-generated solutions for algebra problems. Similarly, number theory problems involving Diophantine equations or integer-valued functional equations exhibit the same issue. Furthermore, we found that the \textit{Proof by Example} fallacy occurs more frequently in algebra, combinatorics, and number theory problems compared to geometry. This trend arises because many problems in these three areas can be framed as proving statements of the form \( Q(x) \), where \( x \) belongs to a specific domain defined by the problem. In such cases, LLMs frequently attempt to verify the proposition \( Q \) by evaluating selected examples from its domain rather than constructing a general proof, thus resulting in the \textit{Proof by Example} fallacy.

\section{Automatic Evaluation Results}
Verification of a candidate solution to a problem is generally considered an easier task than solving the problem itself. Consequently, a common strategy for training LLMs for reasoning is based on a generator-verifier schema. Within this schema, a generator produces candidate solutions and a verifier evaluates these candidates and serves as a reward model. The verifier can be hard coded (for example, check only the correctness of final answers), trained to evaluate solutions produced by LLMs, or a powerful LLM acting as a judge \citep{plaat2024reasoninglargelanguagemodels}. A pertinent question in this context is whether state-of-the-art models can reliably distinguish incorrect solutions from correct solutions. We approach this question through two complementary analyses.

\begin{enumerate}
    \item Do LLMs label correct solutions as correct more frequently than incorrect ones, e.g., those that contain one or more reasoning fallacies?
    \item When presented with pairs of solutions consisting of a correct and an incorrect solution for each problem, do LLMs accurately choose the correct solution?
\end{enumerate}

To investigate the first question, we collected all the problems from our evaluation data set for which the LLM-generated solutions.produced solutions were incorrect. The correct solutions for each of these problems were obtained from the Art of Problem Solving (AoPS) website\footnote{\url{https://artofproblemsolving.com/community/c3223\_imo\_shortlist}}. We then asked the LLMs to analyze each solution and explicitly request a final judgment of either \textit{correct} or \textit{incorrect}. Although this verification task can be complicated due to solutions that may partially satisfy correctness criteria, the solution pairs we chose include a correct solution and an incorrect solution that includes an obvious fallacy. The detailed prompts used in this section can be found in the appendix. 

\begin{table}[t]
\centering
\renewcommand{\arraystretch}{1.2}
\setlength{\tabcolsep}{4pt}
\begin{tabular}{lcc}
\hline
\textbf{Model} & \textbf{Real Solutions Correct (\%)} & \textbf{Wrong Solutions Correct (\%)} \\
\hline
DeepSeek                 & 48 & 43 \\
Gemini 2.0               & 52 & 50 \\
o1                       & 31 & 39 \\
o1-mini                  & 36 & 45 \\
o3-mini                  & 26 & 31 \\
\hline
\end{tabular}
\caption{Percentage of correct and incorrect solutions identified as \textit{correct} by different LLMs during verification. The results illustrate the LLMs' difficulty in accurately distinguishing genuinely correct solutions from clearly incorrect ones containing explicit fallacies.}
\label{tab:solution-verification-results}
\end{table}

As shown in Table \ref{tab:solution-verification-results}, DeepSeek and Gemini 2.0 produce responses of \textit{correct} and \textit{wrong} with nearly equal frequency for both genuinely correct solutions and incorrect LLM-generated solutions. Interestingly, the likelihood of identifying a truly correct solution as \textit{correct} is even lower for the models o1, o1-mini, and o3-mini. These results demonstrate that the models are not suitable for use as judges, as they cannot reliably distinguish genuine solutions from obviously incorrect ones.

To investigate the second question, we applied a similar methodology. For each problem, we presented the models with pairs consisting of a correct solution and an incorrect LLM-generated solution. We then asked the models to identify the correct solution after careful analysis, explicitly informing them that exactly one solution was correct and the other incorrect.

\begin{table}[htbp]
\centering
\renewcommand{\arraystretch}{1.2}
\setlength{\tabcolsep}{10pt}
\begin{tabular}{lc}
\hline
\textbf{Model} & \textbf{Accuracy (\%)} \\
\hline
DeepSeek                 & 57 \\
Gemini 2.0 & 49 \\
o1                       & 50 \\
o1-mini                  & 46 \\
o3-mini                  & 52 \\
\hline
\end{tabular}
\caption{Accuracy of various LLMs in identifying the correct solution when presented with pairs consisting of one correct solution and one incorrect solution generated by an LLM.}
\label{tab:solution-pair-accuracy}
\end{table}

As shown by results in in Table \ref{tab:solution-pair-accuracy}, models o1, o1-mini and Gemini 2.0 perform at or below random to distinguish correct from incorrect solutions. Only DeepSeek and o3-mini perform modestly better than chance, outperforming random selection by 7\% and 2\%, respectively. These results indicate that the evaluated models currently have limited effectiveness as verifiers for challenging tasks such as IMO shortlist-level problems.

\section{Conclusion}
Our evaluation of frontier LLMs on Olympiad-level mathematics revealed significant shortcomings in their ability to produce logically rigorous proofs and engage in genuine mathematical reasoning. Models such as OpenAI's o1, o1-mini, o3-mini, DeepSeek R1, indicate a reliance on heuristic shortcuts rather than authentic reasoning processes. Additionally, these LLMs demonstrated limited capability in effectively verifying solutions, performing at or near random levels when distinguishing correct proofs from clearly incorrect ones.

These findings emphasize two critical areas for improvement. First, there is a clear need to develop more sophisticated benchmarks and evaluation methods to assess the correctness of reasoning and not merely the correctness of final answers. Second, relying solely on the correctness of final answers or utilizing more powerful LLMs as judges is inadequate for assessing the reasoning abilities of LLMs; improved training schemas specifically designed to address the logical rigor of solutions are essential for advancing the reasoning abilities of LLMs.

\bibliography{conference}
\bibliographystyle{conference}

\appendix
\section{Appendix}
\subsection{Illustrative Examples for Common LLM Fallacies}
In this section, we present illustrative examples and comprehensive explanations for each fallacy, facilitating clearer interpretation and deeper conceptual understanding.

\begin{center}
    \begin{tcolorbox}[
    colframe=blue!50!black,         
    colback=blue!3!white,          
    coltitle=white,                
    colbacktitle=modernblue,       
    title=\textbf{Example: Proof by Example},        
    boxrule=0.75pt,
    arc=5pt,
    before skip=17pt,
    after skip=17pt,
    breakable
    ]
    \textbf{Problem:}  
    Prove that $3^{2n} - 1$ is divisible by 8 for all integers $n \geq 1$.

    \vspace{1em}
    \hrule 
    \vspace{1em}

    \textbf{Fallacious Solution:}  
    To "prove" this statement, we can test the initial cases:  

    For \( n = 1 \),  
    \[
    3^2 - 1 = 8, \quad \text{which is divisible by 8.}
    \]

    For \( n = 2 \),  
    \[
    3^4 - 1 = 81 - 1 = 80, \quad \text{which is divisible by 8.}
    \]

    For \( n = 3 \),  
    \[
    3^6 - 1 = 729 - 1 = 728, \quad \text{which is divisible by 8.}
    \]

    Hence, we "proved" that $3^{2n} - 1$ is divisible by 8 for all integers $n \geq 1$.

    \end{tcolorbox}
\end{center}

This argument is not a valid proof by induction, as it lacks the necessary inductive step. Without this inductive step, verifying a few initial cases does not guarantee the statement is true for all \( n \). While the statement itself happens to be true in this example, the method of proof is fallacious.

\vspace{0.5em}

    \begin{tcolorbox}[
    colframe=blue!50!black,         
    colback=blue!3!white,          
    coltitle=white,                
    colbacktitle=modernblue,       
    title=\textbf{Example: Proposal Without Verification:},        
    boxrule=0.75pt,
    arc=5pt,
    before skip=7pt,
    after skip=7pt,
    breakable
    ]

    \textbf{Problem:}  
Two players, Alice and Bob, take turns choosing a number from the set \( \{1, 2, \dots, n\} \), such that the sum of the chosen numbers never exceeds \( 2n \). The player unable to make a valid move loses. Prove that for all \( n \geq 1 \), there exists a winning strategy for one of the players.
    \vspace{1em}
    \hrule 
    \vspace{1em}

    \textbf{Fallacious Solution:}  
    Alice always has a winning strategy. She should begin by picking the largest available number, and on each subsequent turn, she should pick numbers that ensure Bob cannot reach the limit of \( 2n \). By doing this, she can control the game and guarantee her victory. This approach is clearly optimal because by keeping Bob under pressure, Alice can always force him into a losing position."

    \end{tcolorbox}

While this strategy seems intuitive, the proposer does not provide any proof that it guarantees a win for Alice in all cases. For example, there is no justification that Alice’s strategy prevents Bob from achieving the same goal or that Bob has no counter-strategy. Furthermore, no rigorous argument is provided to show that this method works for all values of \( n \). A proper solution would involve analyzing the game mathematically, determining conditions under which Alice or Bob has a winning strategy, and proving these results using tools like invariants or backward induction. Without this reasoning, the proposal remains incomplete and exemplifies the fallacy.

\begin{center}
    \begin{tcolorbox}[
    colframe=blue!50!black,         
    colback=blue!3!white,          
    coltitle=white,                
    colbacktitle=modernblue,       
    title=\textbf{Example: Inventing Wrong Facts}, 
    boxrule=0.75pt,
    arc=5pt,
    before skip=17pt,
    after skip=17pt,
    breakable
    ]

    \textbf{Problem:}  
    Prove that the harmonic series 
    \[
    \sum_{n=1}^\infty \frac{1}{n}
    \]
    diverges.

    \vspace{1em}
    \hrule 
    \vspace{1em}

    \textbf{Fallacious Solution:}  
    By the \textit{Harmonic Square Root Theorem}, the partial sum of the harmonic series up to \( n \) terms is given by:
    \[
    H_n = 2\sqrt{n} + \gamma + \mathcal{O}\!\left(\frac{1}{\sqrt{n}}\right),
    \]
    where \( \gamma \) is the Euler-Mascheroni constant. As \( n \to \infty \), the dominant term \( 2\sqrt{n} \) grows without bound, while the constant \( \gamma \) and error term \( \mathcal{O}\!\left(\frac{1}{\sqrt{n}}\right) \) become negligible. Thus, \( \lim_{n \to \infty} H_n = \infty \), proving divergence of the harmonic series.

    \end{tcolorbox}
\end{center}

The \textit{Harmonic Square Root Theorem} is non-existent. While \( H_n \sim \ln(n) + \gamma \) is a valid asymptotic approximation, claiming \( H_n \propto \sqrt{n} \) is blatantly false.

    \vspace{0.5em}

    \begin{center}
    \begin{tcolorbox}[
    colframe=blue!50!black,         
    colback=blue!3!white,          
    coltitle=white,                
    colbacktitle=modernblue,       
    title=\textbf{Example: Begging the Question}, 
    boxrule=0.75pt,
    arc=5pt,
    before skip=7pt,
    after skip=7pt,
    breakable
    ]

    \textbf{Problem:}  
    Prove that the number \( \sqrt{2} \) is irrational.

    \vspace{1em}
    \hrule 
    \vspace{1em}

    \textbf{Fallacious Solution:}  
    Assume \( \sqrt{2} \) is irrational. By definition, an irrational number cannot be expressed as a fraction \( \frac{a}{b} \), where \( a \) and \( b \) are integers with \( b \neq 0 \). Since \( \sqrt{2} \) cannot be written as \( \frac{a}{b} \), it follows that \( \sqrt{2} \) is irrational. 

    \end{tcolorbox}
\end{center}
The argument begins by assuming the conclusion "Assume \( \sqrt{2} \) is irrational". It then uses the definition of irrationality to "prove" the same claim, creating a logical loop. The conclusion ("\( \sqrt{2} \) is irrational") is restated verbatim as the premise, offering no actual proof.

\vspace{0.5em}

\begin{center}
\begin{tcolorbox}[
colframe=blue!50!black,         
colback=blue!3!white,          
coltitle=white,                
colbacktitle=modernblue,       
title=\textbf{Example: Solution by Trial and Error}, 
boxrule=0.75pt,
arc=5pt,
   before skip=7pt,
    after skip=7pt,
    breakable
]

\textbf{Problem:}  
Find all functions \( f : \mathbb{R} \to \mathbb{R} \) such that  
\[ f(x^2 - y^2) = x f(x) - y f(y), \quad \forall x, y \in \mathbb{R}. \]

\vspace{1em}
\hrule 
\vspace{1em}

\textbf{Fallacious Solution:}  

Let us try some simple forms of \( f(x) \):  
    Suppose \( f(x) = 0 \).  
    Substituting into the functional equation:  
    \[
    f(x^2 - y^2) = 0 \quad \text{and} \quad x f(x) - y f(y) = 0.
    \]
    This satisfies the equation, so \( f(x) = 0 \) is a solution.

    Now suppose \( f(x) = x \).  
    Substituting \( f(x) = x \) into the equation:  
    \[
    f(x^2 - y^2) = x^2 - y^2 \quad \text{and} \quad x f(x) - y f(y) = x^2 - y^2.
    \]
    This also satisfies the equation, so \( f(x) = x \) is another solution.

    Finally, consider \( f(x) = kx \) for some constant \( k \). Substituting:  
    \[
    f(x^2 - y^2) = k(x^2 - y^2) \quad \text{and} \quad x f(x) - y f(y) = k(x^2 - y^2).
    \]

This works for any \( k \).  So the solutions are \( f(x) = 0 \) and \( f(x) = kx \) for any constant \( k \).

\end{tcolorbox}
\end{center}
This solution seems to rely on educated guesses to reach the correct answer but doesn’t explain why other functions fail to satisfy the given functional equation.

\begin{center}
\begin{tcolorbox}[
colframe=blue!50!black,         
colback=blue!3!white,          
coltitle=white,                
colbacktitle=modernblue,       
title=\textbf{Example: Calculation Mistakes}, 
boxrule=0.75pt,
arc=5pt,
before skip=7pt,
after skip=7pt,
breakable
]

\textbf{Problem:}  
Let \( x, y, z > 0 \) satisfy \( x + y + z = 6 \). Find the maximum value of \( xyz\).

\vspace{1em}
\hrule 
\vspace{1em}

\textbf{Fallacious Solution:}  

Using the \textit{AM-GM inequality}, we know:  
\[
\frac{x + y + z}{3} \geq \sqrt[3]{xyz}.
\]

Substituting \( x + y + z = 6 \), we get:  
\[
\frac{6}{3} \geq \sqrt[3]{xyz}.
\]

Simplifying:  
\[
2 \geq \sqrt[3]{xyz}.
\]

Now, squaring both sides (instead of cubing):  
\[
4 \geq xyz.
\]

Thus, we conclude that \( xyz \leq 4 \).

\end{tcolorbox}

\vspace{1em}

\end{center}
The critical error lies in Step 4, where the inequality was squared instead of being cubed. Since the inequality involves a cube root \( \sqrt[3]{xyz} \), the proper operation should have been cubing both sides:  $(2)^3 = 8.$This mistake leads to the incorrect conclusion \( xyz \leq 4 \), whereas the correct bound is \( xyz \leq 27 \). The solution becomes invalid because squaring distorts the inequality, making the result overly restrictive.

\subsection{Real Examples for Common LLM Fallacies}

\begin{center}
\begin{tcolorbox}[
colframe=blue!50!black,         
colback=blue!3!white,          
coltitle=white,                
colbacktitle=modernblue,       
title=\textbf{Example: Proof by Example}, 
boxrule=0.75pt,
arc=5pt,
before skip=7pt,
after skip=7pt,
breakable
]

\textbf{Problem:}  
    On a flat plane in Camelot, King Arthur builds a labyrinth $\mathfrak{L}$ consisting of $n$ walls, each of which is an infinite straight line. No two walls are parallel, and no three walls have a common point. Merlin then paints one side of each wall entirely red and the other side entirely blue. At the intersection of two walls there are four corners: two diagonally opposite corners where a red side and a blue side meet, one corner where two red sides meet, and one corner where two blue sides meet. At each such intersection, there is a two-way door connecting the two diagonally opposite corners at which sides of different colours meet. After Merlin paints the walls, Morgana then places some knights in the labyrinth. The knights can walk through doors, but cannot walk through walls.

Let $k(\mathfrak{L})$ be the largest number $k$ such that, no matter how Merlin paints the labyrinth $\mathfrak{L},$ Morgana can always place at least $k$ knights such that no two of them can ever meet. For each $n,$ what are all possible values for $k(\mathfrak{L}),$ where $\mathfrak{L}$ is a labyrinth with $n$ walls?
\vspace{1em}
\hrule 
\vspace{1em}

\textbf{Fallacious Solution:}  \\
  For each integer \( n \geq 1 \), the labyrinth \( \mathfrak{L} \) constructed by King Arthur consists of \( n \) non-parallel, infinitely extending walls with no three walls intersecting at a single point. Merlin paints each wall with one side red and the other blue. The intersections of these walls create doors that connect specific regions within the labyrinth.

\textbf{Understanding \( k(\mathfrak{L}) \):}

The value \( k(\mathfrak{L}) \) represents the largest number \( k \) such that, regardless of how Merlin paints the labyrinth, Morgana can always place at least \( k \) knights in the labyrinth where no two knights can meet through the doors.

\color{red}
\textbf{Analyzing Small Cases:}

\textbf{1. For \( n = 1 \):}
\begin{itemize}
    \item There is only one wall, dividing the plane into two regions.
    \item Since there are no intersections, there are no doors.
    \item Morgana can place one knight in each region, so \( k(\mathfrak{L}) = 2 \).
\end{itemize}

\textbf{2. For \( n = 2 \):} 
\begin{itemize}
    \item Two intersecting walls create four regions.
    \item There is one door connecting two diagonally opposite regions.
    \item Morgana can place three knights such that no two are connected by the door, so \( k(\mathfrak{L}) = 3 \).
\end{itemize}

\textbf{3. For \( n = 3 \):}

\begin{itemize}
    \item Three walls intersecting pairwise create seven regions.
    \item There are three doors, each connecting two regions.
    \item Morgana can ensure that four knights are placed without any two being connected by a door, so \( k(\mathfrak{L}) = 4 \).
\end{itemize}
\color{black}

\textbf{Generalizing the Pattern:}

\color{red}
Observing the pattern from the small cases, we see that:
\begin{itemize}
    \item \textbf{For each \( n \), \( k(\mathfrak{L}) = n + 1 \).}
\end{itemize}
\color{black}

This means that no matter how Merlin paints the labyrinth, Morgana can always place \( n + 1 \) knights such that no two knights can meet through the doors. This result stems from the combinatorial structure of the labyrinth and the way doors are created at intersections.

\textbf{Conclusion:}

For any number of walls \( n \), the largest guaranteed number of knights \( k(\mathfrak{L}) \) that Morgana can place without any two being able to meet is always \( n + 1 \).

\textbf{Answer:}

For every number \( n \) of walls, the maximal \( k \) is \( n + 1 \). Thus, each \( k(\mathfrak{L}) \) equals \( n + 1 \).

\end{tcolorbox}
\end{center}

The proof claims that \( k(\mathfrak{L}) = n+1\) by testing it for small $n$s and then it generalizes this claim for all natural $n$s without justification.

\begin{center}
\begin{tcolorbox}[
colframe=blue!50!black,         
colback=blue!3!white,          
coltitle=white,                
colbacktitle=modernblue,       
title=\textbf{Example: Proposal Without Verification}, 
boxrule=0.75pt,
arc=5pt,
before skip=7pt,
after skip=7pt,
breakable
]

\textbf{Problem:}  
 A social network has $2019$ users, some pairs of whom are friends. Whenever user $A$ is friends with user $B$, user $B$ is also friends with user $A$. Events of the following kind may happen repeatedly, one at a time:
Three users $A$, $B$, and $C$ such that $A$ is friends with both $B$ and $C$, but $B$ and $C$ are not friends, change their friendship statuses such that $B$ and $C$ are now friends, but $A$ is no longer friends with $B$, and no longer friends with $C$. All other friendship statuses are unchanged.
Initially, $1010$ users have $1009$ friends each, and $1009$ users have $1010$ friends each. Prove that there exists a sequence of such events after which each user is friends with at most one other user.

\vspace{1em}
\hrule 
\vspace{1em}

\textbf{Fallacious Solution:}  \\
To solve this problem, we need to demonstrate that it's possible to transform the given social network into a state where each user is friends with at most one other user using the specified operations.

\textbf{Initial Setup}

\begin{itemize}
    \item \textbf{Number of Users:} 2019
    \item \textbf{Friendship Structure:}
    \begin{itemize}
        \item 1010 users have 1009 friends each.
        \item 1009 users have 1010 friends each.
    \end{itemize}
    \item \textbf{Total Friendships Initially:}
    \[
    1010 \times 1009 + 1009 \times 1010 = 2 \times 1010 \times 1009
    \]
    This counts each friendship twice (once for each user in the pair), so the actual number of unique friendships is:
    \[
    \text{Total Edges} = 1010 \times 1009
    \]
\end{itemize}

\textbf{Objective}

Transform the network so that each user has \textbf{at most one friend}. In graph theory terms, we aim to reach a state where the graph is a \textbf{matching}, meaning it consists of disjoint edges with no overlaps.

\textbf{Operation Details}

The operation allows us to:

\begin{itemize}
    \item \textbf{Select Three Users:} \( A, B, C \) such that:
    \begin{itemize}
        \item \( A \) is friends with both \( B \) and \( C \).
        \item \( B \) and \( C \) are \textbf{not} friends.
    \end{itemize}
    \item \textbf{Modify Friendships:}
    \begin{itemize}
        \item \textbf{Add:} Friendship between \( B \) and \( C \).
        \item \textbf{Remove:} Friendships between \( A \) and \( B \), and between \( A \) and \( C \).
    \end{itemize}
\end{itemize}

\textbf{Effect of Each Operation}

\begin{itemize}
    \item \textbf{Total Friendships Decrease:} Each operation removes two friendships and adds one, resulting in a net decrease of one friendship.
    \item \textbf{Degree Changes:}
    \begin{itemize}
        \item \( A \)'s degree decreases by 2.
        \item \( B \) and \( C \)'s degrees remain unchanged since one friendship is removed and another is added.
    \end{itemize}
\end{itemize}

\textbf{Strategy to Achieve the Objective}

\begin{itemize}
    \item \textbf{Reducing Degrees:}
    \begin{itemize}
        \item Repeatedly apply the operation to reduce the degrees of users with more than one friend.
        \item Each operation ensures that a user's degree decreases, moving towards the desired state.
    \end{itemize}
    \item \textbf{Termination Condition:}
    \begin{itemize}
        \item The process continues until no user has more than one friend.
        \item If a user still has more than one friend, the operation can still be applied because there will always be at least two friends who aren't connected (given the initial high degree and the specific structure of operations).
    \end{itemize}
    \item \textbf{Final State:}
    \begin{itemize}
        \item The network will consist of disjoint pairs of friends (a matching) and possibly some users with no friends.
        \item This ensures that no user has more than one friend.
    \end{itemize}
\end{itemize}

\textbf{Conclusion}

\color{red}
Through systematic application of the specified operation, we can decrementally reduce the degrees of all users with more than one friend. Given the initial conditions and the nature of the operation, it's guaranteed that we can reach a state where each user has at most one friend.
\color{black}

\textbf{Final Answer:}  
After performing suitable operations, the network can be transformed so no user has more than one friend. Thus, such a sequence exists.

\end{tcolorbox}
\end{center}

As we can see, the model has proposed a sequence but doesn't do any mathematical verification of its correctness.

\begin{center}
\begin{tcolorbox}[
colframe=blue!50!black,         
colback=blue!3!white,          
coltitle=white,                
colbacktitle=modernblue,       
title=\textbf{Example: Inventing Wrong Facts}, 
boxrule=0.75pt,
arc=5pt,
before skip=7pt,
after skip=7pt,
breakable
]

\textbf{Problem:}
  You are given a set of $n$ blocks, each weighing at least $1$; their total weight is $2n$. Prove that for every real number $r$ with $0 \leq r \leq 2n-2$ you can choose a subset of the blocks whose total weight is at least $r$ but at most $r + 2$

\vspace{1em}
\hrule 
\vspace{1em}

\textbf{Fallacious Solution:}  \\
To prove that for any real number \( r \) with \( 0 \leq r \leq 2n-2 \), there exists a subset of blocks whose total weight is at least \( r \) but at most \( r + 2 \), we can proceed as follows:

\textbf{Key Observations:}
\begin{enumerate}
    \item \textbf{Block Weights and Total Weight:}
    \begin{itemize}
        \item Each block has a weight of at least 1.
        \item The total weight of all \( n \) blocks is \( 2n \).
    \end{itemize}

    \item \textbf{Subset Sums:}
    \begin{itemize}
        \item The possible subset sums range from 0 (empty set) to \( 2n \) (all blocks).
        \item Since each block has a weight of at least 1, any two distinct subset sums differ by at least 1.
    \end{itemize}

    \item \textbf{Covering the Interval \([0, 2n-2]\):}
    \begin{itemize}
        \item We need to ensure that for any \( r \) in \([0, 2n-2]\), there exists a subset sum \( s \) such that \( r \leq s \leq r + 2 \).
    \end{itemize}
\end{enumerate}

\textbf{Proof Strategy:}

\textbf{Sequential Coverage}
\begin{itemize}

    \item List all possible subset sums in increasing order: \( s_0 = 0, s_1, s_2, \ldots, s_m = 2n \).

    \color{red}
    \item Each subsequent sum satisfies \( s_{i+1} - s_i \geq 1 \) because each block has a weight of at least 1.
    \color{red}
\end{itemize}

\textbf{Interval Overlap}
\begin{itemize}
    \item Consider intervals of the form \([s_i, s_i + 2]\) for each subset sum \( s_i \).
    \item Since \( s_{i+1} \geq s_i + 1 \), the next interval \([s_{i+1}, s_{i+1} + 2]\) starts no more than 1 unit after the start of the previous interval.
    \item Therefore, the union of these intervals \([s_i, s_i + 2]\) covers the entire range \([0, 2n]\) without gaps larger than 2.
\end{itemize}

\textbf{Ensuring Coverage for Any \( r \)}
\begin{itemize}
    \item For any \( r \) in \([0, 2n-2]\), there exists some subset sum \( s_i \) such that \( s_i \leq r \leq s_i + 2 \).
    \item This guarantees that \( r \) lies within the interval \([s_i, s_i + 2]\), meaning there exists a subset with a sum in \([r, r + 2]\).
\end{itemize}

\textbf{Conclusion}
By sequentially covering the range \([0, 2n]\) with overlapping intervals of length 2 derived from the subset sums, we ensure that for any \( r \) within the specified range, there exists a suitable subset whose total weight satisfies \( r \leq \text{sum}(S) \leq r + 2 \).

\textbf{Answer}
By ordering all subset sums and noting that each differs by at least 1, the intervals \([s, s + 2]\) cover \([0, 2n-2]\). Thus, for any \( r \) between \( 0 \) and \( 2n-2 \), some subset has total weight at least \( r \) and at most \( r + 2 \).

\end{tcolorbox}
\end{center}

The model invents a wrong fact/observation and builds its argument around it. The main fact that is used for the proof is obviously not true. Consider $\{1, 1.5, 2, 3.5\}$.

\begin{center}
\begin{tcolorbox}[
colframe=blue!50!black,         
colback=blue!3!white,          
coltitle=white,                
colbacktitle=modernblue,       
title=\textbf{Example: Begging the Question}, 
boxrule=0.75pt,
arc=5pt,
before skip=7pt,
after skip=7pt,
breakable
]

\textbf{Problem:}  
 A polynomial $P(x, y, z)$ in three variables with real coefficients satisfies the identities

$$P(x, y, z)=P(x, y, xy-z)=P(x, zx-y, z)=P(yz-x, y, z).$$
Prove that there exists a polynomial $F(t)$ in one variable such that

$$P(x,y,z)=F(x^2+y^2+z^2-xyz).$$
\vspace{1em}
\hrule 
\vspace{1em}

\textbf{Fallacious Solution:}  \\
 To prove that there exists a polynomial \( F(t) \) in one variable such that
\[
P(x, y, z) = F(x^2 + y^2 + z^2 - xyz),
\]
we analyze the given conditions and the structure of the polynomial \( P(x, y, z) \).

\textbf{Step 1: Define the Invariant}
Let’s define the invariant:
\[
t = x^2 + y^2 + z^2 - xyz.
\]
Our goal is to show that \( t \) remains unchanged under the transformations provided in the problem statement.

\textbf{Step 2: Verify Invariance Under Given Transformations}
We need to verify that \( t \) is invariant under the following transformations:

\textbf{1. First Transformation}
\[
P(x, y, z) = P(x, y, xy - z).
\]
Compute \( t(x, y, xy - z) \):
\[
t = x^2 + y^2 + (xy - z)^2 - x y (xy - z).
\]
Expanding this:
\[
t = x^2 + y^2 + x^2y^2 - 2xyz + z^2 - x^2y^2 + xyz = x^2 + y^2 + z^2 - xyz.
\]
Thus, \( t \) remains unchanged.

\textbf{2. Second Transformation}
\[
P(x, y, z) = P(x, zx - y, z).
\]
Compute \( t(x, zx - y, z) \):
\[
t = x^2 + (zx - y)^2 + z^2 - x(zx - y)z.
\]
Expanding this:
\[
t = x^2 + z^2x^2 - 2xyz + y^2 + z^2 - x^2z^2 + xyz = x^2 + y^2 + z^2 - xyz.
\]
Again, \( t \) remains unchanged.

\textbf{3. Third Transformation}
\[
P(x, y, z) = P(yz - x, y, z).
\]
Compute \( t(yz - x, y, z) \):
\[
t = (yz - x)^2 + y^2 + z^2 - (yz - x)yz.
\]
Expanding this:
\[
t = y^2z^2 - 2xyz + x^2 + y^2 + z^2 - y^2z^2 + xyz = x^2 + y^2 + z^2 - xyz.
\]
Once more, \( t \) remains unchanged.
\color{red}
\\
\\
\textbf{Step 3: Conclude the Existence of \( F(t) \)}
Since \( t = x^2 + y^2 + z^2 - xyz \) is invariant under all the given transformations, and \( P(x, y, z) \) remains unchanged under these transformations, it follows that \( P \) depends only on \( t \). 

Therefore, there exists a polynomial \( F(t) \) such that:
\[
P(x, y, z) = F(t) = F(x^2 + y^2 + z^2 - xyz).
\]
\color{black}
\textbf{Conclusion}
Any polynomial \( P(x, y, z) \) satisfying the given identities must be a polynomial in the single variable \( t = x^2 + y^2 + z^2 - xyz \).

\textbf{Answer}
Because \( x^2 + y^2 + z^2 - xyz \) is unchanged by the given equalities, \( P \) depends only on it. Thus, there is an \( F \) with:
\[
P(x, y, z) = F(x^2 + y^2 + z^2 - xyz).
\]

\end{tcolorbox}
\end{center}

The generated solution begs the question. The fact $P(x, y, z) = F(x^2 + y^2 + z^2 - xyz).$ is the problem statement itself, and it needs to be proven. 

\begin{center}
\begin{tcolorbox}[
colframe=blue!50!black,         
colback=blue!3!white,          
coltitle=white,                
colbacktitle=modernblue,       
title=\textbf{Example: Solution by Trial and Error}, 
boxrule=0.75pt,
arc=5pt,
before skip=7pt,
after skip=7pt,
breakable
]

\textbf{Problem:}  
Let $\mathbb{Z}$ be the set of integers. Determine all functions $f: \mathbb{Z} \rightarrow \mathbb{Z}$ such that, for all integers $a$ and $b$,$$f(2a)+2f(b)=f(f(a+b)).$$Proposed by Liam Baker, South Africa
\vspace{1em}
\hrule 
\vspace{1em}

\textbf{Fallacious Solution:}  \\
To determine all functions \( f: \mathbb{Z} \rightarrow \mathbb{Z} \) that satisfy the functional equation
\[
f(2a) + 2f(b) = f(f(a + b)) \quad \text{for all integers } a \text{ and } b,
\]
let's explore possible forms of \( f \).

\color{red}
\textbf{Step 1: Assume a Linear Function}\\
\color{black}

First, consider that \( f \) might be a linear function of the form:
\[
f(n) = k n + c,
\]
where \( k \) and \( c \) are constants in \( \mathbb{Z} \).

Substituting into the functional equation:
\[
f(2a) + 2f(b) = k(2a) + c + 2(kb + c) = 2k a + c + 2k b + 2c = 2k(a + b) + 3c.
\]
On the other hand:
\[
f(f(a + b)) = f(k(a + b) + c) = k(k(a + b) + c) + c = k^2(a + b) + k c + c.
\]
Setting the two expressions equal gives:
\[
2k(a + b) + 3c = k^2(a + b) + k c + c.
\]
For this to hold for all integers \( a \) and \( b \), the coefficients must match:
\[
2k = k^2 \quad \text{and} \quad 3c = k c + c.
\]
Solving the first equation:
\[
k^2 - 2k = 0 \implies k(k - 2) = 0 \implies k = 0 \text{ or } k = 2.
\]
\textbf{Case 1:} \( k = 0 \)
\[
3c = c \implies 2c = 0 \implies c = 0.
\]
So, \( f(n) = 0 \) is a solution.
\\
\\
\textbf{Case 2:} \( k = 2 \)
\[
3c = 3c \quad \text{(always true)}.
\]
Here, \( c \) can be any integer. Thus, \( f(n) = 2n + c \) is a solution for any integer \( c \).

\textbf{Step 2: Verify Non-Linear Solutions}
Assuming \( f \) is not linear leads to more complex conditions that are difficult to satisfy uniformly across all integers. Hence, no non-linear solutions appear to exist beyond the linear forms already identified.

\textbf{Conclusion}
All solutions to the functional equation are either the constant zero function or affine linear functions of the form \( f(n) = 2n + c \), where \( c \) is an integer constant.
\\
\textbf{Final Answer:} All integer functions equal to twice \( n \) plus a fixed constant. In other words, every \( f \) with \( f(n) = 2n + c \) for some integer \( c \).

\end{tcolorbox}
\end{center}
The generated proof only tests the problem's equation for a special class and concludes the result. This is an example of a solution by trial and error fallacy.

\begin{center}
\begin{tcolorbox}[
colframe=blue!50!black,         
colback=blue!3!white,          
coltitle=white,                
colbacktitle=modernblue,       
title=\textbf{Example: Calculation Mistakes}, 
boxrule=0.75pt,
arc=5pt,
before skip=7pt,
after skip=7pt,
breakable
]

\textbf{Problem:}  
Determine all integers $ n\geq 2$ having the following property: for any integers $a_1,a_2,\ldots, a_n$ whose sum is not divisible by $n$, there exists an index $1 \leq i \leq n$ such that none of the numbers$$a_i,a_i+a_{i+1},\ldots,a_i+a_{i+1}+\ldots+a_{i+n-1}$$is divisible by $n$. Here, we let $a_i=a_{i-n}$ when $i >n$.

\vspace{1em}
\hrule 
\vspace{1em}

\textbf{Fallacious Solution:}  

\textbf{We will show that among the integers \( n \ge 2 \), the property}

\begin{quote}
\textit{“For any integers \(a_1,a_2,\dots,a_n\) whose sum is not divisible by \(n\), there is some index \(1\le i\le n\) so that none of the numbers}
\[
a_i,\; a_i+a_{i+1},\; \dots,\; a_i+a_{i+1}+\cdots+a_{i+n-1}
\]
\textit{is divisible by \(n\)”}
\end{quote}

\noindent holds if and only if \(n = 2\).

In other words, if \(n \ge 3\) then one may construct a counterexample sequence, while for \(n = 2\) the property always holds.

In what follows, we describe a method to translate the condition into one involving a cyclic sequence of residues and then show that counterexamples exist for every \(n \ge 3\), but not for \(n = 2\).

\textbf{Reformulating the Problem}

For a given sequence \(a_1,\dots,a_n\), define the cumulative sums
\[
S_k = a_1 + a_2 + \cdots + a_k \quad \text{for } k=1,\dots,n, \qquad S_0 = 0.
\]
All congruences below are modulo \(n\).

Note that for any index \(i\) (with indices taken cyclically modulo \(n\), i.e., \(a_{j+n} = a_j\)), the partial sums
\[
a_i,\; a_i+a_{i+1},\; \dots,\; a_i+a_{i+1}+\cdots+a_{i+n-1}
\]
can be written as
\[
S_i - S_{i-1},\; S_{i+1} - S_{i-1},\; \dots,\; S_{i+n-1} - S_{i-1}.
\]
Thus, none of these numbers is divisible by \(n\) if and only if
\[
S_{i-1} \not\equiv S_{i-1+j} \quad \text{for } j = 1,2,\dots,n.
\]
In other words, in the list \(S_0, S_1, \dots, S_n\) (indices modulo \(n+1\)), we seek an index \(t\) (with \(t = i-1\)) such that the residue \(S_t\) is isolated—that is, no other \(S_k\) is congruent to \(S_t\) modulo \(n\).

Since \(\mathbb{Z}/n\mathbb{Z}\) has \(n\) elements, but we are choosing \(n+1\) values \(S_0,\dots,S_n\), the pigeonhole principle guarantees that at least one residue is repeated. However, for the property to hold, we only need that at least one \(S_t\) be unique among the \(S_k\).

Thus, the problem reduces to the following:

\begin{quote}
\textit{For which integers \(n \ge 2\) is it true that for every choice of integers \(a_1,\dots,a_n\) with \(S_n \not\equiv 0\) there exists \(t \in \{0,1,\dots,n\}\) such that}
\[
S_t \not\equiv S_k \quad \text{for all } k \ne t\, ?
\]
\end{quote}

If such a \(t\) always exists, then setting \(i = t+1\) ensures that all partial sums starting at \(a_i\) are nonzero modulo \(n\).

\textbf{Counterexamples for \(n \ge 3\)}

We now describe how to construct counterexamples when \(n \ge 3\).

\begin{itemize}
    \item It is possible to construct a sequence \(S_0, S_1, \dots, S_n\) in \(\mathbb{Z}/n\mathbb{Z}\), with \(S_0 = 0\) and \(S_n \not\equiv 0\), such that every residue appears at least twice.
    \item In such a configuration, no \(S_t\) is unique, so the desired property fails.
\end{itemize}

\color{red} One simple way to construct such a sequence is to alternate between two values (e.g., \(0\) and \(x\)), for example:
\[
S_0 = 0,\quad S_1 = 0,\quad S_2 = x,\quad S_3 = 0,\quad S_4 = x,\quad \dots,\quad S_n = x
\]
where \(x\not\equiv 0 \mod n\). Then define
\[
a_k = S_k - S_{k-1},\quad \text{for } k=1,\dots,n.
\]
This ensures that \(a_1 + \cdots + a_n = S_n \not\equiv 0\), and because every residue among \(S_0, \dots, S_n\) appears at least twice, the property fails.

Thus, for every \(n \ge 3\), one can construct a sequence where no index \(i\) satisfies the given condition.
\color{black}

\textbf{The Case \(n = 2\)}

Now consider \(n = 2\). In \(\mathbb{Z}/2\mathbb{Z}\), there are only two residues: \(0\) and \(1\).

\begin{itemize}
    \item The sequence \(S_0, S_1, S_2\) has 3 elements in a 2-element set.
    \item By the pigeonhole principle, at least one residue appears twice.
    \item Therefore, the third must appear only once. So there exists a unique \(S_t\).
\end{itemize}

Hence, the property \textit{must} hold for \(n = 2\).

A concrete check:
\begin{itemize}
    \item If \(a_1\) is odd and \(a_2\) is even, then \(a_1 \equiv 1\), \(a_1+a_2 \equiv 1\).
    \item If \(a_1\) is even and \(a_2\) is odd, then taking \(i = 2\), \(a_2 \equiv 1\), and \(a_2+a_1 \equiv 1\).
\end{itemize}

So in both cases, the two partial sums are nonzero modulo \(2\), confirming the property.

\textbf{Conclusion}

\begin{itemize}
    \item For all \(n \ge 3\), one may construct a sequence \(a_1,\dots,a_n\) such that for every \(i\), at least one of the cyclic partial sums is divisible by \(n\).
    \item For \(n = 2\), such a counterexample is impossible, and the property always holds.
\end{itemize}

Therefore, the only integer \(n \ge 2\) for which the property holds is:
\[
\boxed{2}.
\]

\textbf{Final Answer:} The property holds if and only if \(n = 2\).

\end{tcolorbox}
\end{center}

If $n$ is odd, then $S_n$ will be $0$, which means the sum of $a_1 +a_2+\dots+a_n$ is divisible by $n$, and this contradicts the problem statement. The model has made a calculation mistake in this case and failed to realize that $S_n \neq x$ for odd $n$s

\clearpage

\subsection{Evaluation Prompts}

\begin{center}
\begin{tcolorbox}[
    colframe=teal!50!black,         
    colback=teal!5!white,           
    coltitle=white,                
    colbacktitle=teal!70!black,     
    title=\textbf{Prompt: Solution Evaluation Prompt},
    boxrule=0.75pt,
    arc=5pt,
    before skip=17pt,
    after skip=17pt,
    breakable
]
You are an expert mathematician tasked with rigorously evaluating the correctness of a proposed solution to a mathematical problem, specifically in the style of the International Mathematical Olympiad (IMO).

\vspace{1em}
Below, you will receive:  

\begin{enumerate}
    \item \textbf{The Problem Statement:} A description of the problem to be solved.  
    \item \textbf{The Candidate Solution:} A step-by-step explanation of a proposed solution to the problem.
\end{enumerate}

\vspace{1em}
Your evaluation should include:

\begin{itemize}
    \item \textbf{Logical Analysis:} Carefully check each step for logical consistency, correctness of mathematical reasoning, and appropriateness of methods used.
    \item \textbf{Identification of Errors:} Clearly pinpoint and describe any mathematical errors, flawed logic, unjustified assumptions, or missing steps.
    \item \textbf{Clarification:} Briefly explain why each identified issue is problematic, referring explicitly to mathematical principles or concepts involved.
\end{itemize}

Provide your evaluation strictly adhering to the following structured response format:

\vspace{1em}
\hrule
\vspace{1em}

\textbf{Analysis:} \\
\{Provide a thorough, step-by-step analysis of the candidate solution. Clearly indicate if and where logical fallacies, incorrect assumptions, or mathematical inaccuracies occur. If the solution is fully correct, justify your reasoning thoroughly.\}

\vspace{1em}
\textbf{Final Answer:} \\
Clearly indicate the result of your analysis using exactly one of the following formats:
\[
\boxed{\text{Correct}} \quad \text{or} \quad \boxed{\text{Wrong}}
\]

\end{tcolorbox}
\end{center}

\begin{center}
\begin{tcolorbox}[
    colframe=teal!50!black,         
    colback=teal!5!white,           
    coltitle=white,                
    colbacktitle=teal!70!black,     
    title=\textbf{Prompt: Solution Comparison Prompt},
    boxrule=0.75pt,
    arc=5pt,
    before skip=17pt,
    after skip=17pt,
    breakable
]
I will present a problem along with two candidate solutions. Your task is to evaluate the solutions and determine which one is correct. Note that \textbf{one solution is always correct and one solution is always wrong}.

\vspace{1em}
The primary criterion for your judgment is the \textbf{correctness of the arguments} provided in each solution. 

\vspace{1em}
Please respond using the following template:

\vspace{1em}
\hrule
\vspace{1em}

\textbf{Analysis:} \\
\{Provide your analysis of the solutions here, discussing the strengths and weaknesses of their arguments.\}

\vspace{1em}
\textbf{Final Answer:} \\
\\
\{
\boxed{\text{Solution 1}} \quad \text{or} \quad \boxed{\text{Solution 2}}
\}

\end{tcolorbox}
\end{center}

\subsection{Additional Notes}
To mitigate recency bias, the solutions provided by the LLM and the real solutions were randomly permuted during the comparative analysis experiment. Consistent outcomes were observed across various prompts, indicating robustness in the experimental setup. Furthermore, our investigation revealed that automated rating rubrics, such as the LLM-as-a-judge approach, yielded similar comparative results. Notably, frontier LLMs did not assign significantly higher ratings to correct solutions compared to incorrect ones. Due to constraints in space, the present discussion is intentionally restricted to the binary evaluation scenario.
 
\end{document}